\documentclass[10pt, a4paper]{article}
\usepackage{lrec}
\usepackage{multibib}
\newcites{languageresource}{Language Resources}
\usepackage{graphicx}
\usepackage{tabularx}
\usepackage{soul}
\usepackage{titlesec}
\titleformat{\section}{\normalfont\large\bf\center}{\thesection.}{1em}{}
\titleformat{\subsection}{\normalfont\SmallTitleFont\bf\raggedright}{\thesubsection.}{1em}{}
\titleformat{\subsubsection}{\normalfont\normalsize\bf\raggedright}{\thesubsubsection.}{1em}{}
\renewcommand\thesection{\arabic{section}}
\renewcommand\thesubsection{\thesection.\arabic{subsection}}
\renewcommand\thesubsubsection{\thesubsection.\arabic{subsubsection}}
\usepackage{epstopdf}
\usepackage[utf8]{inputenc}

\usepackage{hyperref}
\usepackage{xstring}

\usepackage{color}

\usepackage{rotating}
\usepackage{tikz}
\usepackage{todonotes}
\usepackage{amssymb}
\usepackage{amsmath}
\usepackage{booktabs}
\usepackage{multirow}
\usepackage{algorithm}
\usepackage{pifont} 
\usepackage{pifont}% http://ctan.org/pkg/pifont
\title{LinCE: A Centralized Benchmark for Linguistic Code-switching Evaluation}
\name{Gustavo Aguilar, Sudipta Kar, Thamar Solorio}

\address{University of Houston\\
         Department of Computer Science \\
         \texttt{\{gaguilaralas, skar3, tsolorio\}@uh.edu}\\
         }
\abstract{
Recent trends in NLP research have raised an interest in linguistic code-switching (CS); modern approaches have been proposed to solve a wide range of NLP tasks on multiple language pairs. Unfortunately, these proposed methods are hardly generalizable to different code-switched languages. In addition, it is unclear whether a model architecture is applicable for a different task while still being compatible with the code-switching setting. This is mainly because of the lack of a centralized benchmark and the sparse corpora that researchers employ based on their specific needs and interests. To facilitate research in this direction, we propose a centralized benchmark for \textbf{Lin}guistic \textbf{C}ode-switching \textbf{E}valuation (\textbf{LinCE}) that combines ten corpora covering four different code-switched language pairs (i.e., Spanish-English, Nepali-English, Hindi-English, and Modern Standard Arabic-Egyptian Arabic) and four tasks (i.e., language identification, named entity recognition, part-of-speech tagging, and sentiment analysis). As part of the benchmark centralization effort, we provide an online platform
at \texttt{ritual.uh.edu/lince},
where researchers can submit their results while comparing with others in real-time. In addition, we provide the scores of different popular models, including LSTM, ELMo, and multilingual BERT so that the NLP community can compare against state-of-the-art systems. LinCE is a continuous effort, and we will expand it with more low-resource languages and tasks.
\\ \newline \Keywords{code-switching, code-mixing, multilingualism, benchmark} }

\begin{document}

\maketitleabstract

\section{Introduction}

% MOTIVATION:
% - The lack of consistent evaluation in code-switching.
% - Models are not tested across tasks to guarantee code-switching generalization
% - More datasets are popping up but we do not have a standard benchmark to compare across methods

Linguistic code-switching\footnote{We use code-switching and code-mixing equivalently.} (CS) is the multilingual phenomenon that happens when speakers alternate languages within the same sentence or utterance. 
% The linguistic code-switching\footnote{We use code-switching and code-mixing equivalently.} (CS)
During the last decade, the CS phenomenon has attracted more research interest from the NLP community. Many researchers have proposed novel methods to handle code-switched data, showing improvements on core NLP tasks such as language identification (LID), named entity recognition (NER), and part-of-speech (POS) tagging. However, many of these approaches are usually evaluated on a few language pairs and a specific domain, and it is not clear whether these models are exclusive to such scenarios or they can generalize to other tasks, domains, and language pairs.

% Moreover, the lack of a centralized code-switching benchmark has led to a slow process of comparison in which researchers have to replicate previous methods to report scores on different datasets. 
Moreover, research in code-switching currently has a slow process of comparison in which researchers have to replicate previous methods to report scores on different datasets. 
Furthermore, choosing the best-published model for benchmarking purposes is not an easy task either. 
These problems exist mainly because 1) there is no official benchmark for general code-switching evaluation that allows direct comparisons across multiple tasks, and 2) methods are usually not comprehensively evaluated across datasets with different language pairs.

To overcome these problems, we propose a centralized \textbf{Lin}guistic \textbf{C}ode-switching \textbf{E}valuation (\textbf{LinCE}) benchmark. We have consolidated a benchmark from preexisting corpora considering the following aspects: 
1) multiple language pairs from high- and low-resource languages with a reasonable range of code-mixing indexes (CMI) \cite{gamback2014measuring}, 
2) typologically-diverse languages\footnote{We also consider the geolocation of such languages to account for places across the world.}, 
3) a variety of NLP tasks including core tasks and downstream applications, and 
4) different code-switching domains from social media platforms. 
LinCE is comprised of four LID datasets, two POS tagging datasets, three NER datasets, and one sentiment analysis (SA) dataset, providing a total of ten datasets (see Table \ref{tab:datasets_vs_tasks}). 
% The language pairs per task are distributed as described in Table \ref{tab:datasets_vs_tasks}.  
% Spanish-English, Nepali-English, Hindi-English, and Modern Standard Arabic-Egyptian Arabic for LID; 
% Hindi-English, and Spanish-English for POS tagging; 
% Hindi-English, Spanish-English, and Modern Standard Arabic-Egyptian Arabic for NER; 
% and Hindi-English and Spanish-English for sentiment analysis. 
Furthermore, an important contribution of LinCE is the new stratification process to provide fair and, in some cases, official splits for the tasks at hand. This required a careful inspection of the original datasets from which we list five major issues (see Section \ref{sec:task:stratification}) and propose new splits for nine out of the ten datasets. 
% To this end, we list the number of issues among the corpora and propose new splits for seven out of the eleven datasets.

\begin{table}[!t]
    \centering
    \resizebox{\linewidth}{!}{
    \begin{tabular}{lcccc}
        \toprule
        \textbf{Language Pair} & \textbf{LID}  & \textbf{POS}  & \textbf{NER}   & \textbf{SA} \\ 
        \midrule
        Spanish-English            & \checkmark & \checkmark & \checkmark & \checkmark \\
        Hindi-English              & \checkmark & \checkmark & \checkmark & - \\
        Nepali-English             & \checkmark &   -        &   -        & - \\
        MS Arabic-Egyptian Arabic  & \checkmark & -          & \checkmark & - \\
        \bottomrule
    \end{tabular}
    }
    \caption{Overview of the LinCE language pairs and tasks.}
    \label{tab:datasets_vs_tasks}
\end{table}

In addition to the LinCE benchmark, we also provide strong baselines using popular models such as LSTMs \cite{lstm:hochreiter:1997}, ELMo \cite{peters-EtAl:2018:N18-1}, and multilingual BERT \cite{devlin2018bert}. In our analysis, we evaluate the results of the best model and describe the outstanding challenges in this benchmark. 
Moreover, LinCE is publicly available at \texttt{ritual.uh.edu/lince}, and we anticipate this benchmark to continue to grow and include new tasks and language pairs as they become available. %will be incorporating more tasks and low-resource language pairs in the near future.
We hope that LinCE motivates future work and accelerates the progress on NLP for code-switched languages.

\section{Related Work}

% CRUCIAL ASPECTS TO COVER:

% - Data-related aspects:
%   - The authors of the datasets used in this benchmark
%   - Previous attempts to evaluate on multiple code-switching datasets
% - Method-related aspects:
%   - Models and methods proposed on 

% Alan Black Survey: sitaram2019survey

Linguistic code-switching has been studied in the context of many NLP tasks \cite{sitaram2019survey}, including 
language identification \cite{solorio-etal-2014-overview,bali-etal-2014-borrowing}, 
part-of-speech tagging \cite{soto-hirschberg-2018-joint,Soto_2017,molina-etal-2016-overview,das-codemixing-contest-icon2016,Solorio:2008:PTE:1613715.1613852}, 
named entity recognition \cite{aguilar-etal-2018-named}, 
parsing \cite{partanen-etal-2018-dependency}, 
sentiment analysis \cite{W15-2902}, 
and question answering \cite{raghavi2015answer,chandu-etal-2018-code}. 
Many code-switching datasets have been made available through the shared-task series 
FIRE \cite{sequiera2015overview,choudhury2014overview,roy2013overview} and 
CALCS \cite{solorio-etal-2014-overview,molina-etal-2016-overview,aguilar-etal-2018-named}, 
which have focused mostly on core NLP tasks. 
Additionally, other researchers have provided datasets for 
dialect recognition \cite{HAMED18.1046}, 
humor detection \cite{khandelwal-etal-2018-humor}, 
sub-word code-switching detection \cite{Mager_2019}, among others. 
Despite the availability and recent growth of datasets, it is still unclear how to compare models across language pairs, domains, and general language processing tasks. 

In the case of language identification (LID) at the token level, researchers have evaluated approaches such as conditional random fields (CRF) with hand-crafted features \cite{al-badrashiny-diab-2016-lili}, LSTM models with word and character embeddings \cite{mave-etal-2018-language,samih-etal-2016-multilingual}, code-mixed word embeddings \cite{pratapa-etal-2018-word}, and transfer learning \cite{aguilar2019english}. While most of these approaches reach over 90\% of accuracy regardless of the language pairs, it is hard to determine which model is the best overall and what the trade-offs are by using one instead of the others. 
Likewise, for part-of-speech (POS) tagging, the community has explored tools that heavily rely on monolingual hand-crafted linguistic information and morphological features \cite{alghamdi2019part}, 
traditional ML techniques (e.g., SVM) with heuristics that exploit monolingual resources \cite{Solorio:2008:PTE:1613715.1613852}, 
combined monolingual taggers including CRF and Random Forest \cite{jamatia-etal-2015-part}, and jointly modeling POS tagging with LID using  recurrent neural networks \cite{soto-hirschberg-2018-joint}. 
Although such approaches are effective on their datasets at hand, they are language-specific and not easy to compare across each other. 

A slightly different trend has been marked in named entity recognition (NER). 
Although the main problem in NER has been the lack of datasets, it is until recently that researchers have provided a few corpora on Hindi-English \cite{singh-etal-2018-twitter}, Spanish-English and Modern Standard Arabic-Egyptian Arabic \cite{aguilar-etal-2018-named}. The participants of the 2018 CALCS competition proposed models based on standard neural NER architectures (e.g., character CNN, followed by a word-based LSTM, and CRF) \cite{geetha-etal-2018-tackling}, including variations with attention \cite{wang-etal-2018-code} and multi-task learning \cite{trivedi-etal-2018-iit}. Additionally, most of the participants exploited publicly available resources such as gazetteers as well as monolingual and multilingual embeddings \cite{winata-etal-2018-bilingual}. 
While the CALCS competition provided datasets on Spanish-English and Modern Standard Arabic-Egyptian Arabic simultaneously, the participants were allowed to provide predictions on one or both competitions. This flexibility left the question open regarding which model was overall the best across language pairs.

Sentiment analysis (SA) on code-switched data has not been explored extensively either. \newcite{W15-2902} provided a Spanish-English polarity annotations for a small section of the 2014 CALCS LID corpus. Their focus was to compare different \textit{bag-of-words} features using L2-regularized logistic regression models often employed in monolingual SA. Concurrent to our work, SemEval-2020 is hosting the first competition for SA in code-switched data, \textit{Task 9: Sentiment Analysis for Code-Mixed Social Media Text} \cite{patwa-etal-2020-sentimix}, which covers Spanish-English and Hindi-English with tweets annotated with both sentiment and language identification labels. We adopt the Spanish-English corpus as part of the LinCE benchmark. 

Although there has been progress in code-switching overall, CS is still lacking advancements in many NLP tasks. 
Additionally, CS tends to advance guided by language-specific challenges, usually providing sparse technologies that may not necessarily be effective for other language pairs.
By gathering different language pairs and tasks into a single benchmark, we expect LinCE to strive for consolidated and steady progress in code-switching research.
% We believe that solid and standard benchmarks strive for steady progress in research, and we expect LinCE to fill that gap in the code-switching community. 

\section{Linguistic Challenges}

\begin{table*}[t!]
    \centering
    \resizebox{\textwidth}{!}{
    \begin{tabular}{lll|rr|rr|rrr} 
        \toprule
        \textbf{Task}
            & \textbf{Corpus} 
            & \textbf{Languages}
            & \textbf{All Posts}  
            & \textbf{All CMI}
            & \textbf{CS Posts} 
            & \textbf{CS CMI} 
            & \textbf{Lang1} 
            & \textbf{Lang2}
            & \textbf{All Tokens}\\
        % \midrule
        % \multicolumn{4}{l}{\textit{Language Identification}} \\
        % SPA: {'cmi_all': 8.288, 'cmi_cm': 21.858, 'all_sents': 32651.0, 'cm_sents': 12380.0}
        % NEP: {'cmi_all': 19.849, 'cmi_cm': 25.751, 'all_sents': 13011.0, 'cm_sents': 10029.0}
        % HIN: {'cmi_all': 10.139, 'cmi_cm': 22.683, 'all_sents': 7421.0, 'cm_sents': 3317.0}
        % MSA: {'cmi_all': 2.818, 'cmi_cm': 23.891, 'all_sents': 11243.0, 'cm_sents': 1326.0}
        \midrule
        \multirow{4}{*}{LID}
            & \newcite{molina-etal-2016-overview}  & SPA-ENG  & 32,651 &  8.29   & 12,380 & 21.86 & 129,065 & 170,793 & 390,953 \\
            & \newcite{solorio-etal-2014-overview} & NEP-ENG  & 13,011 & 19.85   & 10,029 & 25.75 &  59,037 &  78,360 & 188,784 \\
            & \newcite{mave-etal-2018-language}    & HIN-ENG  &  7,421 & 10.14   &  3,317 & 22.68 &  84,752 &  29,958 & 146,722 \\
            & \newcite{molina-etal-2016-overview}  & MSA-EA  & 11,243 &  2.82   &  1,326 & 23.89 & 140,057 &  40,759 & 227,354 \\
        \midrule 
        % \multicolumn{4}{l}{\textit{Part-of-Speech Tagging}} \\
        % \midrule
        % HIN: {'cmi_all': 20.278, 'cmi_cm': 28.035, 'all_sents': 1489.0, 'cm_sents': 1077.0}
        % BGR: {'cmi_all': 24.199, 'cmi_cm': 24.808, 'all_sents': 42911.0, 'cm_sents': 41856.0}
        % SPG: {'cmi_all': 6.69, 'cmi_cm': 25.408, 'all_sents': 695.0, 'cm_sents': 183.0}
        \multirow{2}{*}{POS} 
            & \newcite{singh-etal-2018-twitter}    & HIN-ENG &  1,489  & 20.28  & 1,077   & 28.04 &  12,589 &  9,882 &  33,010 \\
            & \newcite{Soto_2017}                  & SPA-ENG & 42,911  & 24.19  & 41,856  & 24.81 & 178,135 & 92,517 & 333,069 \\
        \midrule
        % \multicolumn{4}{l}{\textit{Named Entity Recognition}} \\
        % \midrule
        % SPA: {'cmi_all': 5.497, 'cmi_cm': 21.156, 'all_sents': 67223.0, 'cm_sents': 17466.0}
        % HIN: {'cmi_all': 19.99, 'cmi_cm': 25.279, 'all_sents': 2079.0, 'cm_sents': 1644.0}
        \multirow{3}{*}{NER} 
            & \newcite{aguilar-etal-2018-named}    & SPA-ENG &  67,223 &   5.49 &  17,466 & 21.16 & 163,824 & 402,923 & 808,663 \\
            & \newcite{singh-etal-2018-language}   & HIN-ENG &   2,079 &  19.99 &   1,644 & 25.28 &  13,860 &  11,391 &  35,374 \\
            & \newcite{aguilar-etal-2018-named}    & MSA-EA &  12,335 &   --   &      -- &    -- &      -- &      -- & 248,478 \\
        \midrule
        % \multicolumn{4}{l}{\textit{Sentiment Analysis}} \\
        % \midrule
        % SPA: {'cmi_all': 20.704, 'cmi_cm': 21.379, 'all_sents': 18789.0, 'all_tokens': 286810, 'cm_sents': 18196.0}
        % HIN: {'cmi_all': 29.787, 'cmi_cm': 29.788, 'all_sents': 20000.0, 'all_tokens': 522601, 'cm_sents': 19999.0}
        % \multirow{2}{*}{SA} 
        \multirow{1}{*}{SA} 
            & \newcite{patwa-etal-2020-sentimix} & SPA-ENG & 18,789 & 20.70 & 18,196 & 21.37 &  65,968 & 144,533 & 286,810 \\
            % & \newcite{patwa-etal-2020-sentimix} & HIN-ENG & 20,000 & 29.78 & 19,999 & 29.78 & 173,385 & 243,019 & 522,601 \\
        \bottomrule
    \end{tabular}
    }
    \caption{The CMI scores and the number of tokens across corpora. \textbf{All Posts} describes the number of posts in the corpora and \textbf{All CMI} is the corresponding CMI scores for such samples. Similarly, \textbf{CS Posts} denotes the number of code-switched posts (excluding monolingual posts) and \textbf{CS CMI} is the corresponding CMI scores for such samples. We also show the number of tokens that belong to the language pairs (\textbf{Lang1}, \textbf{Lang2}) as well as the overall number of tokens (\textbf{All Tokens}), which includes other LID labels beyond the language pairs. English is the \textbf{Lang1} class for English-paired languages; for MSA-EA, Modern Standard Arabic is the \textbf{Lang1} class. We omit the CMI information for the MSA-EA NER corpus because the corpus does not come with language identification labels. }
    % The table shows the CMI scores for the entire corpora (\textbf{All CMI}, along with the number of samples shown in \textbf{All Posts}), as well as the CMI scores for code-switched samples only (\textbf{CS CMI}, accompanied with the number of samples shown in \textbf{CS Posts}). In columns \textbf{Lang1}, \textbf{Lang2} we show the number of tokens that belong to one or the other language, and the overall number of tokens in \textbf{All Tokens}  }
    % \caption{The table shows the CMI scores with respect to the entire corpora (\textbf{All Post} and \textbf{All CMI}) and with respect to the samples that are strictly code-switched (i.e., excluding monolingual posts) across corpora (\textbf{CS Post} and \textbf{CS CMI}).  }
    % \caption{The CMI scores across corpora. The higher the score the more code-switched the text is. We omit the CMI information for the NER MSA-EA corpus since the corpus does not come with language identification labels. English is the \textbf{Lang1} class for English-paired languages; for MSA-EA, Modern Standard Arabic is the \textbf{Lang1} class.}
    \label{tab:dataset_cmi}
\end{table*}

% Linguistic code-switching occurs when multilingual speakers combine multiple languages in the same utterance or sentence, in the case of written language. 
Although code-switching can happen in more than two languages, this benchmark focuses on language pairs only. The frequent alternations between two languages is precisely what makes the automated processing of code-switching data difficult. We quantify such complexity using the CMI index proposed by \newcite{gamback2014measuring} as shown in Table \ref{tab:dataset_cmi}. The higher the CMI index, the more alternations the dataset contains, and hence, the more complex the code-switching behavior is.
% in the corpus.
In addition to the alternation of languages, we briefly describe other linguistic challenges that each specific language pair poses to current NLP systems:

\begin{itemize}
    \item \textbf{Spanish-English (SPA-ENG)}. While English is a Germanic language, a significant number of words from its current vocabulary have been borrowed from Latin and French since the Middle Ages \cite{Tristram1999}. This particular set of words tends to overlap with words from Spanish, a Latin-based language. This overlap increases ambiguity and directly affects systems that rely on character-based approaches, for example, in the case of language identification. Code-switching also appears within the words, often inflecting words by conjugating English verbs using Spanish grammatical rules. This behavior is known as Spanglish \cite{rothman2007linguistic}, and it particularly affects non-contextualized word embeddings as it increases the out-of-vocabulary (OOV) rate.
    
    \item \textbf{Hindi-English (HIN-ENG)}. One of the most challenging aspects of this language pair is the lack of a standardized transliteration system. Speakers transliterate Hindi employing mostly ad-hoc phonological rules to use the English alphabet when writing. Using the same roman alphabet makes code-switching more convenient but the lack of an official standard for transliteration makes it difficult to process with existing resources exclusively available for Hindi with the Devanagari script. Furthermore, although Hindi loosely follows the subject-object-verb (SOV) structure, its flexible word order poses an additional challenge to NLP systems.
    
    \item \textbf{Nepali-English (NEP-ENG)}. Similar to HIN-ENG, Nepali is transliterated using the English alphabet when code-switched with English. This behavior makes Nepali speakers to write driven by arbitrary phonological rules that allow the romanization of Nepali using the English alphabet, which excludes the few monolingual resources available for Nepali. Also, Nepali is a subject-object-verb (SOV) language while English is subject-verb-object (SVO). This grammatical difference intuitively encourages more code-switching points since it is proven that, when code-switching occurs, the languages involved still preserve their grammatical structure \cite{Solorio:2008:PTE:1613715.1613852}, which forces more fine-grained alternations to obey the SOV and SVO structures. In practice, we see a large code-switching rate for Nepali-English captured by the averaged CMI index in Table \ref{tab:dataset_cmi}, being one of the largest scores while having a corpus of middle size.
    
    \item \textbf{Modern Standard Arabic-Egyptian Arabic (MSA-EA)}. 
    Arabic is well known for its diglossia \cite{ferguson1959diglossia}; which combines a number of Arabic dialects with Modern Standard Arabic within the same community. This combination of dialects enables a large occurrence of linguistic code-switching. One of the main challenges with this language pair is that there is a significantly large word overlap while the word meanings can vary depending on the language. Even more, Arabic is a morphologically rich language and it allows multiple word orders, which increases the semantic complexity for NLP systems. Additionally similar to Spanglish, code-switching can occur at the morpheme level, where speakers often add morphological inflections to nouns.
    
\end{itemize}

% \section{Tasks and Datasets}
% \input{tables/dataset_vs_tasks.tex}
\section{Tasks}

LinCE is built upon four tasks and four language pairs to provide a total of eleven datasets. In Sections \ref{sec:task:lid} to \ref{sec:task:sa}, we discuss the datasets used for every task. Then, in Section \ref{sec:task:stratification}, we describe and justify the modifications to nine out of the ten datasets in order to establish official splits that can be adopted for this benchmark. Lastly, in Section \ref{sec:task:evaluation}, we explain the evaluation criteria to rank the leaderboard in the LinCE platform. 
% , including the addition of the validation scores and the online benchmark platform.

% ============================================================
\subsection{Language Identification (LID)}
\label{sec:task:lid}
% ============================================================

\begin{figure}[t!]
    \centering
    \includegraphics[width=\linewidth]{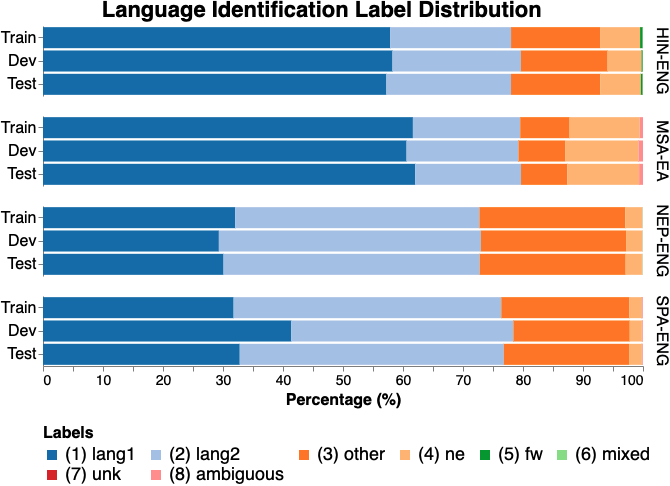}
    \caption{LID label distribution used in LinCE. While HIN-ENG and SPA-ENG have very few tokens for \texttt{unk} and \texttt{fw} ($<$1\%), MSA-EA and NEP-ENG do not have occurrences of such labels. Also, with the exception of MSA-EA, all the partitions are proposed for LinCE as described in Section \ref{sec:task:stratification}. The label scheme is described in Appendix \ref{app:lid-scheme}.}
    \label{fig:lid_label_distribution}
\end{figure}

Handling code-switched data requires to identify the languages involved. The task of language identification (LID) is one of the first steps that validates whether a system can handle code-switched data or not. Correctly classifying the language associated to text units (e.g., words or sub-word tokens) enables to process code-switched text in higher-level applications where general language understanding takes place. 
LinCE uses preexisting datasets for the language identification task. 
Specifically, in this version of LinCE, we focus on the language pairs Spanish-English, Hindi-English, Nepali-English, and Modern Standard Arabic-Egyptian Arabic. 
We briefly explain each corpus below, and for some of them, we propose new splits as explained in the stratification section (Section \ref{sec:task:stratification}). 
Figure \ref{fig:lid_label_distribution} shows the final distribution of the labels across the LID corpora used in LinCE.
Also, these datasets follow the CALCS LID label scheme, which is \texttt{lang1}, \texttt{lang2}, \texttt{mixed} (partially in both languages), \texttt{ambiguous} (either one or the other language), \texttt{fw} (a language different than \texttt{lang1} and \texttt{lang2}), \texttt{ne} (named entities), \texttt{other}, and \texttt{unk} (unrecognizable words). 
More details about the LID label scheme are in Appendix \ref{app:lid-scheme}.

\begin{itemize}
    \item \textbf{SPA-ENG}. We use the Spanish-English corpus from the 2016 CALCS workshop \cite{molina-etal-2016-overview}. This corpus uses Twitter data and it contains 32,651 posts that are comprised of 390,953 tokens. We provide new splits for this corpus because the original splits do not have a similar label distribution and the label \texttt{fw} does not appear in the development set.
	
	\item\noindent\textbf{HIN-ENG}. We use the Hindi-English corpus released by \newcite{mave-etal-2018-language}. This corpus uses Twitter and Facebook data, which have been partly collected and partly re-used from the ICON 2016 competition \cite{sequiera2015pos}. The corpus contains a total 7,421 posts comprised of 146,722 tokens. Also, we proceed with the stratification process on this corpus because the length of the posts were not considered while doing the splits; Twitter has a character length limit in its post, whereas Facebook posts do not have such restriction resulting in significantly longer text. Moreover, the labels \texttt{ambiguous} and \texttt{unk} do not appear in the development set.
	
	\item\noindent\textbf{NEP-ENG}. The Nepali-English corpus comes from the 2014 CALCS workshop \cite{solorio-etal-2014-overview}. This corpus was collected from Twitter and it contains 13,011 posts and 188,784 tokens. We perform a stratification process to provide standard splits for this corpus since the organizers only provided train and test, and the test set does not include any occurrence of the \texttt{ambiguous} class.
	
	\item\noindent\textbf{MSA-EA}. We use the Modern Standard Arabic-Egyptian Arabic corpus from the 2016 CALCS workshop \cite{molina-etal-2016-overview}. This corpus contains Twitter data and it is comprised of 11,243 tweets with 227,354 tokens. 
	Note that there is no occurrence of the labels \texttt{fw} and \texttt{unk} in the entire corpus. We propose new partitions due to the variation across distributions for both the LID labels as well as sentence lengths. 
\end{itemize}

% ============================================================
\subsection{Parts-of-Speech (POS) Tagging}
\label{sec:task:pos}
% ============================================================

\begin{figure}[t!]
    \centering
    \includegraphics[width=\linewidth]{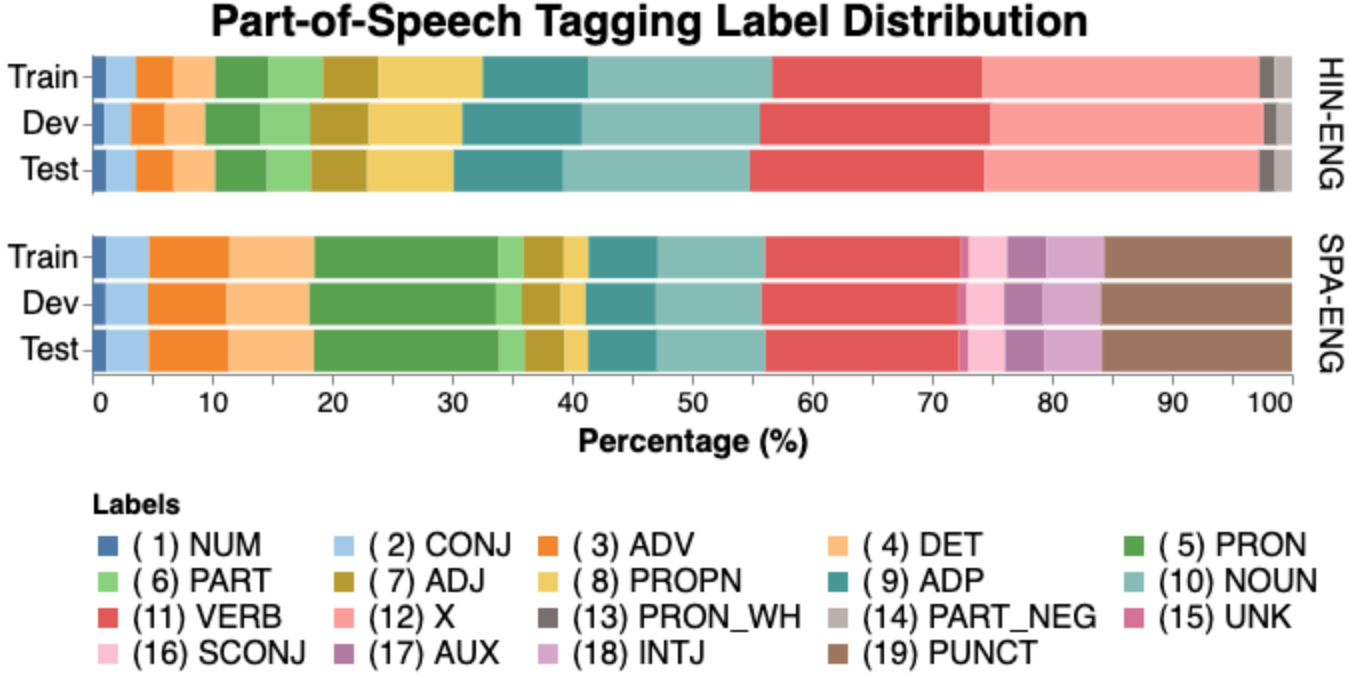}
    \caption{
    POS label distribution used in LinCE. 
    The partitions for both datasets are proposed for LinCE as described in Section \ref{sec:task:stratification}. Note that the labels \texttt{UNK}, \texttt{SCONJ}, \texttt{AUX}, \texttt{INTJ}, and \texttt{PUNCT} only appear in the SPA-ENG corpus, whereas \texttt{PRON\_WH} is unique for HIN-ENG.
    The labels are described in Appendix \ref{app:pos-scheme}}
    \label{fig:pos_label_distribution}
\end{figure}

Part-of-speech (POS) tagging is an important linguistic component that enables more sophisticated syntactic analysis such as constituency and dependency parsing. Code-switched data is not exempted of such analysis. In fact, previous studies have shown that syntax is preserved and compliant with the syntactic rules of the individual languages when code-switching occurs \cite{Solorio:2008:PTE:1613715.1613852}. 
In this benchmark, we consider the language pairs Hindi-English and Spanish-English:

\begin{itemize}
    \item \textbf{HIN-ENG}. \newcite{singh-etal-2018-twitter} provides 1,489 tweets (33,010 tokens) annotated with POS tags and three language IDs (\texttt{hi} for Hindi, \texttt{en} for English, and \texttt{rest} for any other token). 
    The POS tags are annotated using the universal POS tagset proposed by \newcite{petrov-etal-2012-universal} with the addition of two labels: \texttt{PART\_NEG} and \texttt{PRON\_WH}. 
    % Moreover, 30\% of the words are from Hindi, 38\% are from English, and the remaining 32\% are assigned to the \texttt{rest} category.
    % There are 8,796 unique words in the dataset. 
    % Most of the adjectives, nouns, conjunctions, and determiners are English, where Hindi is more frequent for the particles.
    The corpus does not provide training, development, and test splits due to the small number of samples. However, for the purposes of the benchmark, we propose standard splits using the stratification criteria discussed in Section \ref{sec:task:stratification}.
    
    \item \textbf{SPA-ENG}. We use the Miami Bangor corpus with the annotations provided by \newcite{Soto_2017}. The Bangor corpus is composed of bilingual conversations from four speakers with a total of 42,911 utterances and 333,069 tokens. The corpus contains POS tags from the universal POS tagset and LID labels. The LID labels are \texttt{eng} for English, \texttt{spa} for Spanish, \texttt{eng\&spa} for mixed or ambiguous words, and \texttt{UNK} for everything else. Additionally, we proceed with the stratification process to provide the official training, development, and testing sets for this benchmark since the original sets were split by speakers.
    
    % \item \textbf{MSA-EA}. The MSA-EA dataset \cite{solorio-etal-2014-overview} is comprised of code-switched texts collected from Twitter [TWT] and Blog commentaries [COM]. TWT data is divided into two splits for training and testing models. To verify the robustness of models, COM data is chosen as a secondary test set. 
\end{itemize}

% \begin{figure}
%     \centering
%     \fbox{\includegraphics[width=\columnwidth]{figures/precog_pos_lang_bar.png}}
%     \caption{Percentage of words in each language for different POS.}
%     \label{fig:precog_pos_lang_bar}
% \end{figure}

% ============================================================
\subsection{Named Entity Recognition (NER)}
\label{sec:task:ner}
% ============================================================

\begin{figure}[t!]
    \centering
    \includegraphics[width=\linewidth]{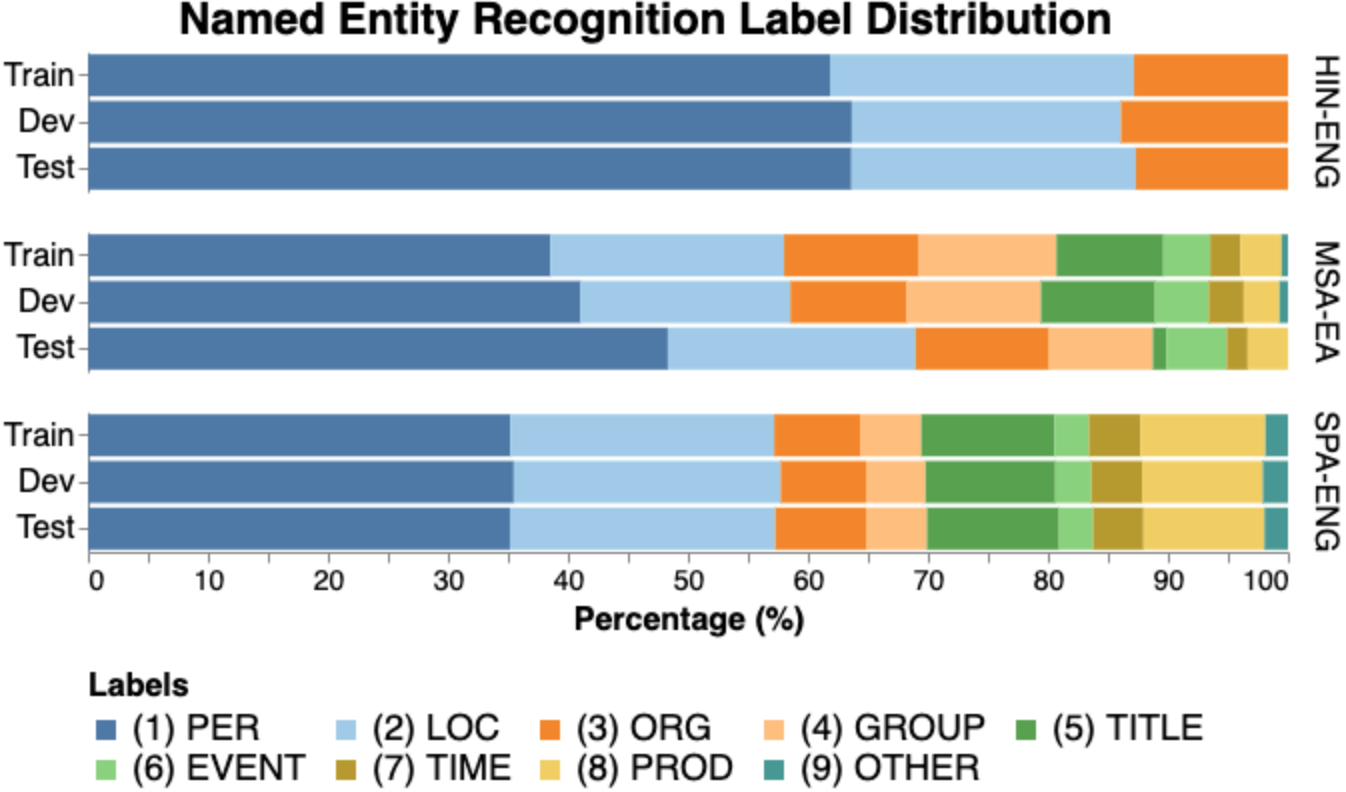}
    \caption{NER label distribution used in LinCE. All the datasets have the BIO scheme, but we only show the entity types for simplicity. 
    Note that HIN-ENG only contains \texttt{PER}, \texttt{LOC}, and \texttt{ORG}.
    Also, with the exception of MSA-EA, all the other partitions are proposed for LinCE as described in Section \ref{sec:task:stratification}. The labels are described in Appendix \ref{app:ner-scheme}.}
    \label{fig:ner_label_distribution}
\end{figure}

Named entity recognition (NER) is another important core NLP task that enables higher-level applications such as question-answering, semantic role labeling, and information extraction. LinCE covers NER for three languages pairs: 
Spanish-English, Modern Standard Arabic-Egyptian Arabic, and Hindi-English.
% The first two corpora were introduced in the CALCS 2018 shared-task \cite{aguilar-etal-2018-named}, and 
% along with the NER labels, we have added the LID classes that were missing in the original release. Moreover, both datasets come from Twitter and they have the same NER labels scheme: \texttt{person}, \texttt{organization}, \texttt{location}, \texttt{group}, \texttt{product}, \texttt{title}, \texttt{event}, \texttt{time}, and \texttt{other}.

\begin{itemize}
    \item \textbf{SPA-ENG}. This corpus was introduced in the 2018 CALCS competition for NER \cite{aguilar-etal-2018-named}, and it contains a total of 67,223 tweets with 808,663 tokens. The labels are \texttt{organization}, \texttt{person}, \texttt{location}, \texttt{group}, \texttt{product}, \texttt{title}, \texttt{event}, \texttt{time}, and \texttt{other}. Along with the NER labels, we have added the LID categories for every token, which follows the CALCS LID scheme. Moreover, we propose new splits for this corpus since the distribution of the NER labels across the splits is not consistent to the one from the full corpus. Additionally, the original development set is significantly small compared to the other splits, only accounting for 832 tweets, and the LID labels were not taken into consideration for the splitting process (e.g., the label \texttt{fw} does not appear in the development set). We provide new splits following the stratified process described in Section \ref{sec:task:stratification}
    
    \item \textbf{MSA-EA}. This corpus was also introduced in the 2018 CALCS competition for NER, following the same entity label scheme as in the SPA-ENG corpus.
    The corpus uses the tweets from the 2016 CALCS LID dataset to form the training and development sets. While the LID labels are available for the training and development splits, the test set was annotated only using the NER labels. Thus, this is the only corpus for which we do not consider the language identification analysis. The corpus contains 12,335 posts and 248,452 tokens. We adopt the splits provided by the organizers during the 2018 CALCS competition.
    
    \item \textbf{HIN-ENG}. This corpus is proposed by \newcite{singh-etal-2018-language}, and it is composed of 2,079 tweets with 35,374 tokens. The dataset has been annotated with both NER and LID labels. The entity labels are \texttt{person}, \texttt{location}, and \texttt{organization}, while the LID labels are \texttt{eng} (English), \texttt{hin} (Hindi), and \texttt{rest} (any other token). This dataset is small, and for that reason, the authors opted to do 5-fold cross validation instead of partitioning the dataset. Nevertheless, for the sake of the benchmark, we split the data using our stratification process that fairly splits the dataset accounting for LID and NER label distributions, as well as the distribution of the tweet lengths.
\end{itemize}

% ============================================================
\subsection{Sentiment Analysis (SA)}
\label{sec:task:sa}
% ============================================================

\begin{figure}[t!]
    \centering
    \includegraphics[width=\linewidth]{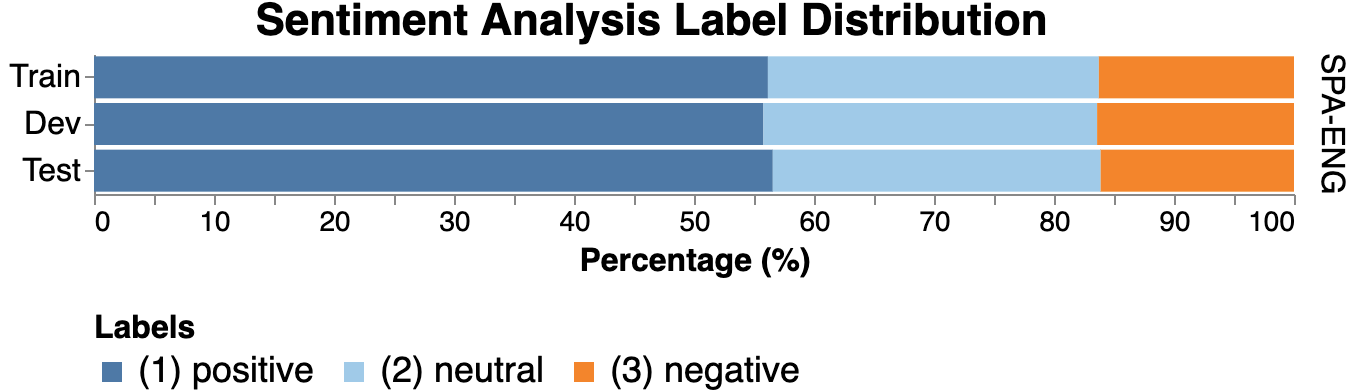}
    \caption{Label distribution of the sentiment analysis corpus used in LinCE. Note that this distribution differs from the origial dataset.
    % We use the same distribution as in the SentiMix competition. 
    }
    \label{fig:sa_label_distribution}
\end{figure}

% In addition to the previous core NLP tasks, we incorporate sentiment analysis as a high-level NLP application. 
We choose sentiment analysis as our fourth benchmark task to incorporate a high-level NLP application in contrast to the previous core NLP tasks.
We use the Spanish-English corpus provided in the SentiMix competition \cite{patwa-etal-2020-sentimix}. %, which covers the language pairs Hindi-English and Spanish-English. 
The organizers reduce the monolingual posts, increasing the number code-switched instances. 
Table \ref{tab:dataset_cmi} shows that this language pair is the second highest scores on the ``All CMI'' column for the sentiment analysis task. 
The task requires to predict one of the sentiments \texttt{positive}, \texttt{negative}, or \texttt{neutral} for every post. 
Additionally, this corpus is annotated with LID labels at the token level, following the CALCS LID scheme, and it contains 18,789 tweets comprised of 286,810 tokens. 
We propose new partitions for this dataset to correct the label distribution from the original splits.
% and HIN-ENG only has \texttt{english}, \texttt{hindi} and \texttt{other}.
% For HIN-ENG, they provide 20,000 tweets in total with 522,601 tokens, whereas SPA-ENG 
% We use the splits released by the organizers.

% ============================================================
\subsection{Stratification}
\label{sec:task:stratification}
% ============================================================

\begin{table*}[t!]
    \centering
    \small
    \resizebox{\textwidth}{!}{
    \begin{tabular}{lll|rrr|rrr|rrr} 
        \toprule
        \multirow{2}{*}{\textbf{Tasks}} 
            & \multirow{2}{*}{\textbf{Corpus Authors}} 
            & \multirow{2}{*}{\textbf{Languages}}
            & \multicolumn{3}{c|}{\textbf{Training}} 
            & \multicolumn{3}{c|}{\textbf{Development}}
            & \multicolumn{3}{c}{\textbf{Test}} \\
        % \cline{4-9}
        ~ & ~ & ~
            & \textbf{CMI}  & \textbf{Posts}  & \textbf{Tokens}
            & \textbf{CMI}  & \textbf{Posts}  & \textbf{Tokens}
            & \textbf{CMI}  & \textbf{Posts}  & \textbf{Tokens} \\
        \midrule
        \multirow{4}{*}{LID}
            & \newcite{molina-etal-2016-overview}  & SPA-ENG  &  8.491  &  21,030  & 253,221 &  7.062 & 3,332  & 40,391 &  8.264 &  8,289  & 97,341 \\
            & \newcite{solorio-etal-2014-overview} & NEP-ENG  & 20.322  &   8,451  & 122,952 & 17.079 & 1,332  & 19,273 & 19.754 &  3,228  & 46,559 \\
            & \newcite{mave-etal-2018-language}    & HIN-ENG  & 10.222  &   4,823  &  95,224 & 10.122 &   744  & 15,446 &  9.930 &  1,854  & 36,052 \\
            & \newcite{molina-etal-2016-overview}  & MSA-EA   &  2.567  &  8,464   & 171,872 &  3.185 &  1,116 & 21,978 &  3.849 &  1,663  & 33,504 \\
        \midrule 
        \multirow{2}{*}{POS} 
            & \newcite{singh-etal-2018-twitter}    & HIN-ENG & 21.449 &   1,030 &  22,993 & 15.293 &    160 &  3,476 & 18.910 &     299 &  6,541 \\
            & \newcite{Soto_2017}                  & SPA-ENG & 24.191 &  27,893 & 217,068 & 24.040 &  4,298 & 33,345 & 24.282 &  10,720 & 82,656 \\
        \midrule
        \multirow{3}{*}{NER}  
            & \newcite{aguilar-etal-2018-named}    & SPA-ENG &  5.567 &  33,611 &  404,428 &  4.398 &  10,085 &  122,656 &  5.867 &  23,527 &  281,579 \\
            & \newcite{singh-etal-2018-language}   & HIN-ENG & 20.117 &   1,243 &   21,065 & 19.913 &     314 &    5,364 & 19.733 &     522 &   8,945 \\
            & \newcite{aguilar-etal-2018-named}    & MSA-EA  &  --    &  10,103 &  204,296 &  --    &   1,122 &   22,742 &     -- &   1,110 &  21,414 \\
        \midrule
        \multirow{1}{*}{SA} 
            & \newcite{patwa-etal-2020-sentimix}   & SPA-ENG  & 20.643 & 12,194 & 186,602 & 21.553 & 1,859 & 28,202 & 20.528 & 4,736 & 72,006 \\
            % & \newcite{patwa-etal-2020-sentimix}   & SPA-ENG & 21.843 & 12,002 & 185,376 & 20.529 & 2,998 & 46,164 & 17.234 & 3,789 & 55,270 \\
            % & \newcite{patwa-etal-2020-sentimix}   & HIN-ENG & 29.765 & 14,000 & 365,560 & 29.965 & 3,000 & 78,678 & 29.709 & 3,000 & 78,363\\
        \bottomrule
    \end{tabular}
    }
    \caption{ Final data distribution of the LinCE benchmark. Note that the proposed distribution follows the stratification process described in Section \ref{sec:task:stratification}, which generates partitions that differ from the original datasets. }
    \label{tab:dataset_splits}
\end{table*}

\begin{table}[t!]
    \centering
    \resizebox{\linewidth}{!}{
    \begin{tabular}{lllll}
        \toprule
        \multirow{2}{*}{\textbf{Task}} 
            & \multirow{2}{*}{\textbf{Dataset}} 
            & \multirow{2}{*}{\textbf{Reason}}
            & \multicolumn{2}{l}{\textbf{KL-divergence}} \\
        ~ & ~ & ~
            & \textbf{before} 
            & \textbf{after}   \\
        \midrule
        \multirow{2}{*}{LID}
            & SPA-ENG  & 1, 2       & 0.10586   & 0.00528  \\
            & HIN-ENG  & 1, 2, 3    & 4.64265   & 0.00064  \\
            & NEP-ENG  & 1, 3, 5    & 0.00552   & 0.00059  \\
            & MSA-EA   &            & 0.17737   & 0.00026  \\
        \midrule
        \multirow{2}{*}{POS}
            & SPA-ENG  & 4, 5       & 0.00140   & 0.00005  \\
            & HIN-ENG  & 5          & --        & 0.00133  \\
        \midrule
        \multirow{2}{*}{NER}
            & SPA-ENG  & 1, 2, 4    & 0.00239   & 0.00001  \\
            & HIN-ENG  & 5          & --        & 0.00007  \\ 
        \midrule
        \multirow{1}{*}{SA}
            & SPA-ENG  & 2, 4       & 0.09579   & 0.00002  \\
        \bottomrule
    \end{tabular}
    }
    \caption{The table shows the datasets for which we propose new splits. The column \textit{Reason} provides the reason number according to the aspects listed in Section \ref{sec:task:stratification}. For the KL-divergence columns, we provide the average scores for the original (\textit{before}) and the proposed (\textit{after}) splits. The lower the score, the more similar the splits are to the full corpus distribution.}
    \label{tab:strat_insights}
\end{table}

For nine out of ten datasets,\footnote{We did not partition the NER MSA-EA dataset because it does not have LID labels, which is essential to keep the code-switching behavior balanced accross splits in our stratification process.} we propose new splits that in our view lead to a more appropriate evaluation (see Table \ref{tab:dataset_splits} for a high-level distribution). We provide new splits for datasets where we found at least one of the following issues:
% \footnote{Unfortunately, not all the datasets were available in a timely manner to provide new partitions with our method. This includes the MSA-EA datasets for LID and NER, as well as the sentiment analysis datasets from SentiMix 2020 (concurrent to this work).}
\begin{enumerate}
    \item At least one of the splits does not have one or more classes. %There is no sample at all for a specific class in at least one split. 
    That is, one or more classes from are not evaluated at all in the development or test set.
    \item The distribution of the label set for a given task is substantially different across splits or against the full corpus distribution (i.e., when merging all the splits into a single set). 
    % See the KL-divergence between the original and proposed splits in Table \ref{tab:strat_insights}.
    \item The length of the sentences do not follow a similar distribution across splits or against the full corpus. This is a relevant criteria to consider since length is positively correlated with context, and less or more context can make a huge different for tasks such as NER.
    \item The NER, POS, and SA datasets contain LID labels that were not considered during the time of the stratification process, potentially affecting the balance of code-switching occurrences across the splits.
    % \item The datasets whose tasks are other than LID (i.e., NER, POS, or SA), do not consider the distribution of the language identification labels, only the distribution of the actual task labels.
    \item There is no official split for the training, development and test sets to provide the scope of fair comparison.
\end{enumerate}

For the datasets where we find at least one of these issues (see Table \ref{tab:strat_insights}), we proceed to stratify based on the language identification labels, if available,
% \footnote{Only MSA-EA for NER does not have LID labels.} 
the task-specific labels (i.e., for tasks other than LID), and the lengths of the sentences. Note that providing splits that consider these three factors jointly in the stratification process is not trivial. In fact, in the case of sequence labeling tasks, we may have multiple non-unique labels per sentence, which constraints the ability to draw a distribution similar to the full corpora (e.g., adding a sentence impacts the distribution of different labels occurring in the same sentence). 

To provide splits considering these three criteria, we follow the iterative stratification process proposed by \newcite{sechidis2011stratification}. 
This process targets multi-label data, which is a different scenario for the document and sequence labeling classification datasets used in LinCE. 
To adapt the sequence labels to the multi-label scenario, we treat a post as a single sample that is associated to a group of labels. 
In the case of tasks other than LID, we gather the LID labels with the task-specific labels (i.e., NER, POS, or SA labels) into a single group of unique labels. 
We also incorporate sentence lengths to the label set of a sample by choosing one of three length categories: \texttt{small} ($\le$ 10 tokens), \texttt{medium} (\textgreater 10 and $\le$ 20 tokens), or \texttt{large} (\textgreater 20 tokens). 
For instance, the SPA-ENG NER sample
\begin{quote}
    \centering
    \textit{``$
        \text{LREC}^{\text{event}}_{\text{ne}} ~
        \text{ser\'a}_{\text{lang2}} ~
        \text{hosted} ~
        \text{in} ~
        \text{Marseille}^{\text{location}}_{\text{ne}}
    $''} \\
    \textbf{English:} \textit{``LREC will be hosted in Marseille''}
\end{quote}
has the set of unique labels \{\texttt{lang1}, \texttt{lang2}, \texttt{ne}, \texttt{event}, \texttt{location}, \texttt{O}, \texttt{small}\}, where the first three labels are for LID, the following three labels are for NER, and the last one represents the sentence length (note that the order and the repetitions of the labels do not matter).
Once we have the set of labels associated to a single sample (e.g., a set of LID, POS, and length labels), we can follow the iterative stratification processed used for multi-label classification on our corpus. We have found that this procedure works well in practice; we measure the KL-divergence of the label distributions from each of the splits against the distribution of the full corpus before and after the stratification, and we found that the proposed splits have less divergence (see Table \ref{tab:strat_insights}). While KL-divergence is not often employed to corroborate the distributions of a stratified corpus, we use the divergence score to quantify whether the distribution of the full corpus has been preserved in the proposed splits, and whether the new splits are better distributions than the original splits. The final numbers of sentences and tokens per partition are listed in Table \ref{tab:dataset_splits}. 

\begin{table*}[t]
    \centering
    \resizebox{\textwidth}{!}{
    \begin{tabular}{l|c|cccc|cc|ccc|cc}
    \toprule
     &
     & \multicolumn{4}{c|}{\textbf{LID (Accuracy) }} 
     & \multicolumn{2}{c|}{\textbf{POS (Accuracy)}} 
     & \multicolumn{3}{c|}{\textbf{NER (Micro F1)}} 
     & \multicolumn{1}{c}{\textbf{SA (Accuracy)}} \\
    \textbf{Model} 
        & \textbf{Avg}
        & \textbf{SPA-ENG} & \textbf{HIN-ENG} & \textbf{NEP-ENG} & \textbf{MSA-EA}  % LID
        & \textbf{SPA-ENG} & \textbf{HIN-ENG}                                       % POS
        & \textbf{SPA-ENG} & \textbf{HIN-ENG} & \textbf{MSA-EA}                     % NER
        & \textbf{SPA-ENG} \\ % & \textbf{HIN-ENG} \\                                 % SA
    \midrule
    % \multicolumn{12}{c}{Single Task Training}\\
    % \midrule
    BiLSTM 
        & 73.20 % 71.57
        & 94.16 & 92.34 & 93.29 & 74.26 
        & 94.80 & 81.84 
        & 44.92 & 48.36 & 62.64
        & 45.39 \\ % & 55.26 \\
    ELMo
        & 78.64 % 76.63 
        & 98.12 & 96.21 & 96.19 & 80.54 
        & 96.30 & 88.42 
        & 53.80 & 65.83 & 61.00
        & 49.97 \\ % & 56.56 \\
    ML-BERT
        & 82.93 % \textbf{82.62}
        & 98.53 & 96.44 & 96.57 & 84.14 
        & 97.00 & 89.28 
        & 63.56 & 75.96 & 67.61
        & 60.20 \\ % & 62.23 \\
    % \midrule
    % \multicolumn{12}{c}{Multi Task Training}\\
    % \midrule
    % ELMo$^\dagger$
    %     & --
    %     & -- & -- & -- & -- 
    %     & -- & -- 
    %     & -- & -- & --
    %     & -- & -- \\
    % ML-BERT$^\dagger$
    %     & --
    %     & -- & -- & -- & -- 
    %     & -- & -- 
    %     & -- & -- & --
    %     & -- & -- \\
    \bottomrule
    \end{tabular}}
    \caption{Baseline results on the test set of the LinCE benchmark.}
    \label{tab:test_results}
\end{table*}
% ============================================================
\subsection{Evaluation}
\label{sec:task:evaluation}
% ============================================================

LinCE adopts an evaluation model similar to SemEval, Kaggle, and GLUE \cite{wang-etal-2018-glue}.
A platform will be hosted at \url{ritual.uh.edu/lince} where participants will be able to upload their predictions for the test data on each task.
The platform will score the submissions and publish the results in a public leaderboard for each task. The leaderboard is ranked by the average of the task scores.

\section{Baseline Models}

We propose LinCE to motivate a single model architecture that has good generalization capability across all the proposed tasks. 
To this end, we experiment with the following model architectures that demonstrated superior performance in a wide range of NLP tasks in recent years.

\begin{itemize}
\item \textbf{Bidirectional LSTM}. 
Our simplest baseline is based on LSTM cells \cite{lstm:hochreiter:1997}, which we operate towards both directions of the texts. For the sequence labeling tasks (i.e., LID, POS, and NER), we concatenate the forward and backward hidden representations for each of the input token and use a linear layer to predict the most probable label for each token.
For sentiment analysis, we perform a max-pooling operation over all the hidden representations of the tokens and use that to predict the most probable sentiment class.
For all of the tasks, we represent each token by a randomly initialized word embedding vector that is tuned during the training process.
    
\item \textbf{ELMo}.
Combining character-level convolution and word-level sequence modeling with LSTM, ELMo \cite{peters-EtAl:2018:N18-1} have shown improvement in various NLP tasks acting as a pre-trained language model.
We fine-tune the publicly available pre-trained ELMo models on the proposed tasks by using its high-level word representations to perform sequence labeling (i.e., LID, POS, and NER).
We use the mean of the token representations to predict sentiment labels for the SA task.

\item \textbf{ML-BERT}
Transformer \cite{transformers} based pre-trained language model like BERT \cite{devlin2018bert} have shown impressive generalized performance in a wide range of natural language understanding tasks.
The strength of such model comes from the large amount of parameters tuned on a huge amount of training data from diverse domains.
We develop our third baseline system with the pre-trained BERT model trained on multilingual data from 104 languages.
We add a task-specific prediction layer over BERT and fine-tune the whole model.

\end{itemize}

\subsection{Implementation and Training}
We implement the models using PyTorch\footnote{\url{pytorch.org}} deep learning library.
We also use AllenNLP \cite{gardner2018allennlp} and HuggingFace's Transformers \cite{Wolf2019HuggingFacesTS} to fine-tune pre-trained ELMo and BERT based models, respectively.
While evaluating a single model architecture across different tasks, we keep the whole model architecture exactly the same except the prediction layer.
We train the BiLSTM and ELMo models using SGD ($\eta=0.1$ for BiLSTM and $\eta=0.01$ for ELMo, $\beta=0.9$).
For fine-tuning BERT, we use the AdamW \cite{adamwloshchilov2018decoupled} ($\eta=5e^{-5}, \epsilon=1e^{-8}$).
We use a batch size of 32 for training every model.
We train each model for a maximum of 50 epochs and stop if the validation performance does not improve for 10 epochs.
All of our experimental choices were tuned by observing the performance on the validation sets.

% \paragraph{Model Architecture}

% \paragraph{Pre-Training Methods}
\section{Results and Analysis}

We report our results on the test sets for the LinCE tasks in Table \ref{tab:test_results} and discuss the details here.
Across almost all the tasks we observe superior performance of the pre-trained language models compared to the simple BiLSTM model.
Among the pre-trained language models, ML-BERT demonstrates superior performance in each task for all the available language pairs.
ELMo's performance is very close to ML-BERT in most of the LID and POS tasks ($\approx$1-4\%), but the performance gap is bigger for NER ($\approx$8-10\%) and SA ($\approx$18\%).
The average performance gap between ELMo and BERT is $\approx$6\%.
We suspect that such improvement for ML-BERT against ELMo is powered by its larger parameter set (110M vs 13.6M) and the amount of data used for pre-training (Book Corpus \cite{bookcorpus} and Wikipedia for BERT and Billion Words Benchmark \cite{billionwordbenchmark} for ELMo).
It is also noteworthy to mention that the training data for BERT is document-level data, whereas the Billion Words Benchmark is a sentence-level corpus.
Among the four tasks, NER and SA seem harder compared to LID and POS.
It shows that the involvement of semantic understanding in code-switched texts makes tasks harder compared to syntactic analysis.

We observe that the LID task is harder for MSA-EA compared to the other language pairs.
A possible reason for this is the large overlap between these two languages, which also affected the annotation process \cite{molina-etal-2016-overview}.
% to the point that even annotators have hard time telling whether one word is in one language or the other. This can explain that we are seeing way lower scores in LID for MSA-EA compared to the other three language pairs

% Thus we conclude by saying that, we may need to draw attention for building resources for language pairs that do not include English language and Roman scripts.

% \input{sections/7-analysis.tex}

\section{Conclusion}

In this paper, we motivate the need for a centralized platform to perform evaluation of technology for code-switching data across multiple tasks and language pairs. To this end, we introduce the \textbf{Lin}guistic \textbf{C}ode-switching \textbf{E}valuation (\textbf{LinCE}) benchmark using ten publicly available datasets. 
In addition, we review such datasets and found important issues that undermine the evaluation process 
(e.g., labels not appearing in the test set, or substantially different distributions among splits, etc.). 
Then, we propose new splits using a new stratification technique with up to three criteria (e.g., LID labels, task-specific labels, and sentence lengths). We show the distribution of the full corpus is preserved in the proposed splits used in LinCE, which is not always the case in the original partitions. Additionally, we provide results with strong baselines using state-of-the-art models on monolingual datasets, including BERT and ELMo. Finally, we expect that LinCE will be well-received by the NLP community, and we will keep the platform evolving with the incorporation of more tasks and language pairs in the near future.
\section{Acknowledgements}

This work was supported by the National Science Foundation (NSF) on the grant \#1910192. 
We thank the authors of the individual datasets that agreed to centralize their corpora into a single benchmark for code-switching. 
Also, we thank the undergraduate students at the University of Houston, Jason Ho, Tarun Appannagari, and Dwija Parikh, that helped on the creation of the LinCE benchmark website.

% \bibliographystyle{lrec}
% \section{Bibliographical References}\label{reference}
% \bibliography{lrec2020W-main}
% \section{Language Resource References}\label{reference-language}
% \bibliography{lrec2020W-langresources}
% \bibliographystylelanguageresource{lrec}
% \bibliographylanguageresource{languageresource}

\section{Bibliographical References}
\label{reference}

\bibliographystyle{lrec}
\bibliography{lrec2020W-main}

\begin{thebibliography}{}

\bibitem[\protect\citename{Aguilar and Solorio}2019]{aguilar2019english}
Aguilar, G. and Solorio, T.
\newblock (2019).
\newblock {From English to Code-Switching: Transfer Learning with Strong
  Morphological Clues}.
\newblock {\em arXiv preprint arXiv:1909.05158}.

\bibitem[\protect\citename{Aguilar \bgroup et al.\egroup
  }2018]{aguilar-etal-2018-named}
Aguilar, G., AlGhamdi, F., Soto, V., Diab, M., Hirschberg, J., and Solorio, T.
\newblock (2018).
\newblock {Named Entity Recognition on Code-Switched Data: Overview of the
  CALCS 2018 Shared Task}.
\newblock In {\em {Proceedings of the Third Workshop on Computational
  Approaches to Linguistic Code-Switching}}, pages 138--147, Melbourne,
  Australia, July. Association for Computational Linguistics.

\bibitem[\protect\citename{Al-Badrashiny and
  Diab}2016]{al-badrashiny-diab-2016-lili}
Al-Badrashiny, M. and Diab, M.
\newblock (2016).
\newblock {LILI: A Simple Language Independent Approach for Language
  Identification}.
\newblock In {\em Proceedings of {COLING} 2016, the 26th International
  Conference on Computational Linguistics: Technical Papers}, pages 1211--1219,
  Osaka, Japan, December. The COLING 2016 Organizing Committee.

\bibitem[\protect\citename{AlGhamdi \bgroup et al.\egroup
  }2019]{alghamdi2019part}
AlGhamdi, F., Molina, G., Diab, M., Solorio, T., Hawwari, A., Soto, V., and
  Hirschberg, J.
\newblock (2019).
\newblock {Part-of-Speech Tagging for Code-Switched Data}.
\newblock {\em arXiv preprint arXiv:1909.13006}.

\bibitem[\protect\citename{Bali \bgroup et al.\egroup
  }2014]{bali-etal-2014-borrowing}
Bali, K., Sharma, J., Choudhury, M., and Vyas, Y.
\newblock (2014).
\newblock {{``}{I} am borrowing ya mixing ?'' An Analysis of {E}nglish-{H}indi
  Code Mixing in {F}acebook}.
\newblock In {\em Proceedings of the First Workshop on Computational Approaches
  to Code Switching}, pages 116--126, Doha, Qatar, October. Association for
  Computational Linguistics.

\bibitem[\protect\citename{Chandu \bgroup et al.\egroup
  }2018]{chandu-etal-2018-code}
Chandu, K., Loginova, E., Gupta, V., Genabith, J.~v., Neumann, G., Chinnakotla,
  M., Nyberg, E., and Black, A.~W.
\newblock (2018).
\newblock {Code-Mixed Question Answering Challenge: Crowd-sourcing Data and
  Techniques}.
\newblock In {\em Proceedings of the Third Workshop on Computational Approaches
  to Linguistic Code-Switching}, pages 29--38, Melbourne, Australia, July.
  Association for Computational Linguistics.

\bibitem[\protect\citename{Chelba \bgroup et al.\egroup
  }2013]{billionwordbenchmark}
Chelba, C., Mikolov, T., Schuster, M., Ge, Q., Brants, T., Koehn, P., and
  Robinson, T.
\newblock (2013).
\newblock {One Billion Word Benchmark for Measuring Progress in Statistical
  Language Modeling}.
\newblock Technical report, Google.

\bibitem[\protect\citename{Choudhury \bgroup et al.\egroup
  }2014]{choudhury2014overview}
Choudhury, M., Chittaranjan, G., Gupta, P., and Das, A.
\newblock (2014).
\newblock {Overview of FIRE 2014 Track on Transliterated Search}.
\newblock {\em Proceedings of FIRE}, pages 68--89.

\bibitem[\protect\citename{Das}2016]{das-codemixing-contest-icon2016}
Das, A.
\newblock (2016).
\newblock {Tool contest on POS tagging for code-mixed Indian social media
  (Facebook, Twitter, and Whatsapp) text}.
\newblock retrieved 05-10-2019.

\bibitem[\protect\citename{Devlin \bgroup et al.\egroup }2018]{devlin2018bert}
Devlin, J., Chang, M.-W., Lee, K., and Toutanova, K.
\newblock (2018).
\newblock {BERT: Pre-training of Deep Bidirectional Transformers for Language
  Understanding}.
\newblock {\em arXiv preprint arXiv:1810.04805}.

\bibitem[\protect\citename{Ferguson}1959]{ferguson1959diglossia}
Ferguson, C.~A.
\newblock (1959).
\newblock Diglossia.
\newblock {\em word}, 15(2):325--340.

\bibitem[\protect\citename{Gamb{\"a}ck and Das}2014]{gamback2014measuring}
Gamb{\"a}ck, B. and Das, A.
\newblock (2014).
\newblock {On Measuring the Complexity of Code-mixing}.
\newblock In {\em Proceedings of the 11th International Conference on Natural
  Language Processing, Goa, India}, pages 1--7. Citeseer.

\bibitem[\protect\citename{Gardner \bgroup et al.\egroup
  }2018]{gardner2018allennlp}
Gardner, M., Grus, J., Neumann, M., Tafjord, O., Dasigi, P., Liu, N., Peters,
  M., Schmitz, M., and Zettlemoyer, L.
\newblock (2018).
\newblock {AllenNLP: A Deep Semantic Natural Language Processing Platform}.
\newblock {\em arXiv preprint arXiv:1803.07640}.

\bibitem[\protect\citename{Geetha \bgroup et al.\egroup
  }2018]{geetha-etal-2018-tackling}
Geetha, P., Chandu, K., and Black, A.~W.
\newblock (2018).
\newblock {Tackling Code-Switched {NER}: Participation of {CMU}}.
\newblock In {\em Proceedings of the Third Workshop on Computational Approaches
  to Linguistic Code-Switching}, pages 126--131, Melbourne, Australia, July.
  Association for Computational Linguistics.

\bibitem[\protect\citename{Hamed \bgroup et al.\egroup }2018]{HAMED18.1046}
Hamed, I., Elmahdy, M., and Abdennadher, S.
\newblock (2018).
\newblock {Collection and Analysis of Code-switch Egyptian Arabic-English
  Speech Corpus}.
\newblock In Nicoletta Calzolari~(Conference chair), et~al., editors, {\em
  Proceedings of the Eleventh International Conference on Language Resources
  and Evaluation (LREC 2018)}, Miyazaki, Japan, May 7-12, 2018. European
  Language Resources Association (ELRA).

\bibitem[\protect\citename{Hochreiter and
  Schmidhuber}1997]{lstm:hochreiter:1997}
Hochreiter, S. and Schmidhuber, J.
\newblock (1997).
\newblock {Long Short-Term Memory}.
\newblock {\em Neural computation}, 9(8):1735--1780.

\bibitem[\protect\citename{Jamatia \bgroup et al.\egroup
  }2015]{jamatia-etal-2015-part}
Jamatia, A., Gamb{\"a}ck, B., and Das, A.
\newblock (2015).
\newblock Part-of-speech tagging for code-mixed {E}nglish-{H}indi twitter and
  {F}acebook chat messages.
\newblock In {\em Proceedings of the International Conference Recent Advances
  in Natural Language Processing}, pages 239--248, Hissar, Bulgaria, September.
  INCOMA Ltd. Shoumen, BULGARIA.

\bibitem[\protect\citename{Khandelwal \bgroup et al.\egroup
  }2018]{khandelwal-etal-2018-humor}
Khandelwal, A., Swami, S., Akhtar, S.~S., and Shrivastava, M.
\newblock (2018).
\newblock Humor detection in {E}nglish-{H}indi code-mixed social media content
  : Corpus and baseline system.
\newblock In {\em Proceedings of the Eleventh International Conference on
  Language Resources and Evaluation ({LREC} 2018)}, Miyazaki, Japan, May.
  European Language Resources Association (ELRA).

\bibitem[\protect\citename{Loshchilov and
  Hutter}2019]{adamwloshchilov2018decoupled}
Loshchilov, I. and Hutter, F.
\newblock (2019).
\newblock Decoupled weight decay regularization.
\newblock In {\em International Conference on Learning Representations}.

\bibitem[\protect\citename{Mager \bgroup et al.\egroup }2019]{Mager_2019}
Mager, M., {\c{C}}etino{\~g}lu, {\"O}., and Kann, K.
\newblock (2019).
\newblock Subword-level language identification for intra-word code-switching.
\newblock {\em Proceedings of the 2019 Conference of the North}.

\bibitem[\protect\citename{Mave \bgroup et al.\egroup
  }2018]{mave-etal-2018-language}
Mave, D., Maharjan, S., and Solorio, T.
\newblock (2018).
\newblock Language identification and analysis of code-switched social media
  text.
\newblock In {\em Proceedings of the Third Workshop on Computational Approaches
  to Linguistic Code-Switching}, pages 51--61, Melbourne, Australia, July.
  Association for Computational Linguistics.

\bibitem[\protect\citename{Molina \bgroup et al.\egroup
  }2016]{molina-etal-2016-overview}
Molina, G., AlGhamdi, F., Ghoneim, M., Hawwari, A., Rey-Villamizar, N., Diab,
  M., and Solorio, T.
\newblock (2016).
\newblock {Overview for the Second Shared Task on Language Identification in
  Code-Switched Data}.
\newblock In {\em Proceedings of the Second Workshop on Computational
  Approaches to Code Switching}, pages 40--49, Austin, Texas, November.
  Association for Computational Linguistics.

\bibitem[\protect\citename{Partanen \bgroup et al.\egroup
  }2018]{partanen-etal-2018-dependency}
Partanen, N., Lim, K., Rie{\ss}ler, M., and Poibeau, T.
\newblock (2018).
\newblock Dependency parsing of code-switching data with cross-lingual feature
  representations.
\newblock In {\em Proceedings of the Fourth International Workshop on
  Computational Linguistics of Uralic Languages}, pages 1--17, Helsinki,
  Finland, January. Association for Computational Linguistics.

\bibitem[\protect\citename{Patwa \bgroup et al.\egroup
  }2020]{patwa-etal-2020-sentimix}
Patwa, P., Aguilar, G., Kar, S., Pandey, S., PYKL, S., Garrette, D.,
  Gamb{\"a}ck, B., Chakraborty, T., Solorio, T., and Das, A.
\newblock (2020).
\newblock {SemEval-2020 Sentimix Task 9: Overview of SENTIment Analysis of
  Code-MIXed Tweets}.
\newblock In {\em Proceedings of the 14th International Workshop on Semantic
  Evaluation ({S}em{E}val-2020)}, Barcelona, Spain, September. Association for
  Computational Linguistics.

\bibitem[\protect\citename{Peters \bgroup et al.\egroup
  }2018]{peters-EtAl:2018:N18-1}
Peters, M., Neumann, M., Iyyer, M., Gardner, M., Clark, C., Lee, K., and
  Zettlemoyer, L.
\newblock (2018).
\newblock {Deep Contextualized Word Representations}.
\newblock In {\em Proceedings of the 2018 Conference of the North American
  Chapter of the Association for Computational Linguistics: Human Language
  Technologies, Volume 1 (Long Papers)}, pages 2227--2237, New Orleans,
  Louisiana, June. Association for Computational Linguistics.

\bibitem[\protect\citename{Petrov \bgroup et al.\egroup
  }2012]{petrov-etal-2012-universal}
Petrov, S., Das, D., and McDonald, R.
\newblock (2012).
\newblock A universal part-of-speech tagset.
\newblock In {\em Proceedings of the Eighth International Conference on
  Language Resources and Evaluation ({LREC}'12)}, pages 2089--2096, Istanbul,
  Turkey, May. European Language Resources Association (ELRA).

\bibitem[\protect\citename{Pratapa \bgroup et al.\egroup
  }2018]{pratapa-etal-2018-word}
Pratapa, A., Choudhury, M., and Sitaram, S.
\newblock (2018).
\newblock Word embeddings for code-mixed language processing.
\newblock In {\em Proceedings of the 2018 Conference on Empirical Methods in
  Natural Language Processing}, pages 3067--3072, Brussels, Belgium,
  October-November. Association for Computational Linguistics.

\bibitem[\protect\citename{Raghavi \bgroup et al.\egroup
  }2015]{raghavi2015answer}
Raghavi, K.~C., Chinnakotla, M.~K., and Shrivastava, M.
\newblock (2015).
\newblock Answer ka type kya he?: Learning to classify questions in code-mixed
  language.
\newblock In {\em Proceedings of the 24th International Conference on World
  Wide Web}, pages 853--858. ACM.

\bibitem[\protect\citename{Rothman and Rell}2007]{rothman2007linguistic}
Rothman, J. and Rell, A.~B.
\newblock (2007).
\newblock A linguistic analysis of spanglish: Relating language to identity.
\newblock {\em Linguistics and the Human Sciences}, 1(3):515--536.

\bibitem[\protect\citename{Roy \bgroup et al.\egroup }2013]{roy2013overview}
Roy, R.~S., Choudhury, M., Majumder, P., and Agarwal, K.
\newblock (2013).
\newblock {Overview of the FIRE 2013 Track on Transliterated Search}.
\newblock In {\em Post-Proceedings of the 4th and 5th Workshops of the Forum
  for Information Retrieval Evaluation}, page~4. ACM.

\bibitem[\protect\citename{Samih \bgroup et al.\egroup
  }2016]{samih-etal-2016-multilingual}
Samih, Y., Maharjan, S., Attia, M., Kallmeyer, L., and Solorio, T.
\newblock (2016).
\newblock Multilingual code-switching identification via {LSTM} recurrent
  neural networks.
\newblock In {\em Proceedings of the Second Workshop on Computational
  Approaches to Code Switching}, pages 50--59, Austin, Texas, November.
  Association for Computational Linguistics.

\bibitem[\protect\citename{Sechidis \bgroup et al.\egroup
  }2011]{sechidis2011stratification}
Sechidis, K., Tsoumakas, G., and Vlahavas, I.
\newblock (2011).
\newblock {On the stratification of Multi-label Data}.
\newblock In {\em Joint European Conference on Machine Learning and Knowledge
  Discovery in Databases}, pages 145--158. Springer.

\bibitem[\protect\citename{Sequiera \bgroup et al.\egroup
  }2015a]{sequiera2015pos}
Sequiera, R., Choudhury, M., and Bali, K.
\newblock (2015a).
\newblock {POS Tagging of Hindi-English Code Mixed Text from Social Media: Some
  Machine Learning Experiments}.
\newblock In {\em 2015 Proceedings of International Conference on NLP}. NLPAI,
  December.

\bibitem[\protect\citename{Sequiera \bgroup et al.\egroup
  }2015b]{sequiera2015overview}
Sequiera, R., Choudhury, M., Gupta, P., Rosso, P., Kumar, S., Banerjee, S.,
  Naskar, S.~K., Bandyopadhyay, S., Chittaranjan, G., Das, A., et~al.
\newblock (2015b).
\newblock {Overview of FIRE-2015 Shared Task on Mixed Script Information
  Retrieval}.
\newblock In {\em FIRE Workshops}, volume 1587, pages 19--25.

\bibitem[\protect\citename{Singh \bgroup et al.\egroup
  }2018a]{singh-etal-2018-language}
Singh, K., Sen, I., and Kumaraguru, P.
\newblock (2018a).
\newblock Language identification and named entity recognition in {H}inglish
  code mixed tweets.
\newblock In {\em Proceedings of {ACL} 2018, Student Research Workshop}, pages
  52--58, Melbourne, Australia, July. Association for Computational
  Linguistics.

\bibitem[\protect\citename{Singh \bgroup et al.\egroup
  }2018b]{singh-etal-2018-twitter}
Singh, K., Sen, I., and Kumaraguru, P.
\newblock (2018b).
\newblock A twitter corpus for {H}indi-{E}nglish code mixed {POS} tagging.
\newblock In {\em Proceedings of the Sixth International Workshop on Natural
  Language Processing for Social Media}, pages 12--17, Melbourne, Australia,
  July. Association for Computational Linguistics.

\bibitem[\protect\citename{Sitaram \bgroup et al.\egroup
  }2019]{sitaram2019survey}
Sitaram, S., Chandu, K.~R., Rallabandi, S.~K., and Black, A.~W.
\newblock (2019).
\newblock A survey of code-switched speech and language processing.
\newblock {\em arXiv preprint arXiv:1904.00784}.

\bibitem[\protect\citename{Solorio and
  Liu}2008]{Solorio:2008:PTE:1613715.1613852}
Solorio, T. and Liu, Y.
\newblock (2008).
\newblock Part-of-speech tagging for english-spanish code-switched text.
\newblock In {\em Proceedings of the Conference on Empirical Methods in Natural
  Language Processing}, EMNLP '08, pages 1051--1060, Stroudsburg, PA, USA.
  Association for Computational Linguistics.

\bibitem[\protect\citename{Solorio \bgroup et al.\egroup
  }2014]{solorio-etal-2014-overview}
Solorio, T., Blair, E., Maharjan, S., Bethard, S., Diab, M., Ghoneim, M.,
  Hawwari, A., AlGhamdi, F., Hirschberg, J., Chang, A., and Fung, P.
\newblock (2014).
\newblock Overview for the first shared task on language identification in
  code-switched data.
\newblock In {\em Proceedings of the First Workshop on Computational Approaches
  to Code Switching}, pages 62--72, Doha, Qatar, October. Association for
  Computational Linguistics.

\bibitem[\protect\citename{Soto and Hirschberg}2017]{Soto_2017}
Soto, V. and Hirschberg, J.
\newblock (2017).
\newblock Crowdsourcing universal part-of-speech tags for code-switching.
\newblock {\em Interspeech 2017}, Aug.

\bibitem[\protect\citename{Soto and
  Hirschberg}2018]{soto-hirschberg-2018-joint}
Soto, V. and Hirschberg, J.
\newblock (2018).
\newblock Joint part-of-speech and language {ID} tagging for code-switched
  data.
\newblock In {\em Proceedings of the Third Workshop on Computational Approaches
  to Linguistic Code-Switching}, pages 1--10, Melbourne, Australia, July.
  Association for Computational Linguistics.

\bibitem[\protect\citename{Tristram}1999]{Tristram1999}
Tristram, H. L.~C.
\newblock (1999).
\newblock {\em How Celtic is Standard English?}
\newblock Nauka, St Petersburg, Russia.

\bibitem[\protect\citename{Trivedi \bgroup et al.\egroup
  }2018]{trivedi-etal-2018-iit}
Trivedi, S., Rangwani, H., and Kumar~Singh, A.
\newblock (2018).
\newblock {IIT} ({BHU}) submission for the {ACL} shared task on named entity
  recognition on code-switched data.
\newblock In {\em Proceedings of the Third Workshop on Computational Approaches
  to Linguistic Code-Switching}, pages 148--153, Melbourne, Australia, July.
  Association for Computational Linguistics.

\bibitem[\protect\citename{Vaswani \bgroup et al.\egroup }2017]{transformers}
Vaswani, A., Shazeer, N., Parmar, N., Uszkoreit, J., Jones, L., Gomez, A.~N.,
  Kaiser, L.~u., and Polosukhin, I.
\newblock (2017).
\newblock Attention is all you need.
\newblock In I.~Guyon, et~al., editors, {\em Advances in Neural Information
  Processing Systems 30}, pages 5998--6008. Curran Associates, Inc.

\bibitem[\protect\citename{Vilares \bgroup et al.\egroup }2015]{W15-2902}
Vilares, D., Alonso, M.~A., and G{\'o}mez-Rodr{\'i}guez, C.
\newblock (2015).
\newblock Sentiment analysis on monolingual, multilingual and code-switching
  twitter corpora.
\newblock In {\em Proceedings of the 6th Workshop on Computational Approaches
  to Subjectivity, Sentiment and Social Media Analysis}, pages 2--8.
  Association for Computational Linguistics.

\bibitem[\protect\citename{Wang \bgroup et al.\egroup
  }2018a]{wang-etal-2018-glue}
Wang, A., Singh, A., Michael, J., Hill, F., Levy, O., and Bowman, S.
\newblock (2018a).
\newblock {GLUE}: A multi-task benchmark and analysis platform for natural
  language understanding.
\newblock In {\em Proceedings of the 2018 {EMNLP} Workshop {B}lackbox{NLP}:
  Analyzing and Interpreting Neural Networks for {NLP}}, pages 353--355,
  Brussels, Belgium, November. Association for Computational Linguistics.

\bibitem[\protect\citename{Wang \bgroup et al.\egroup
  }2018b]{wang-etal-2018-code}
Wang, C., Cho, K., and Kiela, D.
\newblock (2018b).
\newblock {Code-Switched Named Entity Recognition with Embedding Attention}.
\newblock In {\em Proceedings of the Third Workshop on Computational Approaches
  to Linguistic Code-Switching}, pages 154--158, Melbourne, Australia, July.
  Association for Computational Linguistics.

\bibitem[\protect\citename{Winata \bgroup et al.\egroup
  }2018]{winata-etal-2018-bilingual}
Winata, G.~I., Wu, C.-S., Madotto, A., and Fung, P.
\newblock (2018).
\newblock Bilingual character representation for efficiently addressing
  out-of-vocabulary words in code-switching named entity recognition.
\newblock In {\em Proceedings of the Third Workshop on Computational Approaches
  to Linguistic Code-Switching}, pages 110--114, Melbourne, Australia, July.
  Association for Computational Linguistics.

\bibitem[\protect\citename{Wolf \bgroup et al.\egroup
  }2019]{Wolf2019HuggingFacesTS}
Wolf, T., Debut, L., Sanh, V., Chaumond, J., Delangue, C., Moi, A., Cistac, P.,
  Rault, T., Louf, R., Funtowicz, M., and Brew, J.
\newblock (2019).
\newblock Huggingface's transformers: State-of-the-art natural language
  processing.
\newblock {\em ArXiv}, abs/1910.03771.

\bibitem[\protect\citename{Zhu \bgroup et al.\egroup }2015]{bookcorpus}
Zhu, Y., Kiros, R., Zemel, R., Salakhutdinov, R., Urtasun, R., Torralba, A.,
  and Fidler, S.
\newblock (2015).
\newblock Aligning books and movies: Towards story-like visual explanations by
  watching movies and reading books.
\newblock In {\em The IEEE International Conference on Computer Vision (ICCV)},
  December.

\end{thebibliography}

% \section{Language Resource References}
% \label{lr:ref}
% \bibliographystylelanguageresource{lrec}
% \bibliographylanguageresource{languageresource}

\clearpage
\appendix
\section*{\centering Appendix for ``LinCE: A Centralized Benchmark for Linguistic Code-switching Evaluation''}
\label{sec:supplemental}

\section{Label Schemes}
\label{app:label-schemes}

\subsection{LID Label Scheme}
\label{app:lid-scheme}

We use the CALCS label scheme for language identification, which contains the following labels:
\begin{itemize}
    \item \texttt{lang1} represents one of the code-switched languages. For English-paired lanaguages, this label is used for English tokens; for MSA-EA, this label is used for Modern Standard Arabic.
    \item \texttt{lang2} represents the other code-switched language. For LinCE, this label could be Spanish, Hindi, Nepali, or Egyptian Arabic, depending on the dataset.
    \item \texttt{mixed} represents a word that is partly in \texttt{lang1} and partly in \texttt{lang2}.
    \item \texttt{ambiguous} is used for words whether it is unclear if they belong to one or the other language.
    \item \texttt{fw} means foreign word, and it is used for words that are in a language different than \texttt{lang1} or \texttt{lang2}.
    \item \texttt{ne} represents the named entity tokens.
    \item \texttt{unk} means unknown, and it is used for tokens whose language is not recognized. 
    \item \texttt{other} captures symbols, punctuation, and emoticons.
\end{itemize}   

\subsection{POS Label Scheme}
\label{app:pos-scheme}

We use the universal part-of-speech (UPOS) tagset proposed by \newcite{petrov-etal-2012-universal} with the addition of \texttt{PRON\_WH} proposed by \newcite{singh-etal-2018-twitter} and \texttt{UNK} proposed by \newcite{Soto_2017}.
\begin{itemize}
    \item \texttt{ADJ} is used for adjectives.
    \item \texttt{ADP} is used for prepositions and postpositions.
    \item \texttt{ADV} is used for adverbs.  
    \item \texttt{AUX} is used for auxiliaries.
    \item \texttt{CONJ} is used for coordinating conjunctions. This is represented by `CCONJ' in the univeral POS tagset. 
    \item \texttt{DET} is used for determiners and articles. 
    \item \texttt{INTJ} is used for interjections.
    \item \texttt{NOUN} is used for nouns.
    \item \texttt{NUM} is used for numerals.
    \item \texttt{PART} is used for particles.
    \item \texttt{PRON} is used for pronouns.
    \item \texttt{PROPN} is used for proper nouns.
    \item \texttt{PUNCT} is used for punctuation marks.
    \item \texttt{SCONJ} is used for subordinating conjunctions. 
    \item \texttt{VERB} is used for verbs.
    \item \texttt{X} for all other categories such as abbreviations or foreign words.
    \item \texttt{PRON\_WH} is used for interrogative pronouns (like where, why, etc.). This extension is employed by \newcite{singh-etal-2018-twitter}.
    \item \texttt{UNK} is used when it is not possible to determine the syntactic category. This extension is employed by \newcite{Soto_2017}.
\end{itemize}

\subsection{NER Label Scheme}
\label{app:ner-scheme}

We use the CALCS label scheme for named entity recognition \cite{aguilar-etal-2018-named}. These labels use the BIO scheme and contain the following entity types:
\begin{itemize}
    \item \texttt{person} for proper nouns or nicknames.
    \item \texttt{organization} for institutions, companies, organizations, or corporations. Not to confuse with products when they have the same name as an organization.
    \item \texttt{location} for physical places that people can visit and that have a unique name. Addresses, facilities, and touristic places are examples of this. 
    \item \texttt{group} for sports teams, music bands, duets, etc. Not to confuse with organization.
    \item \texttt{product} for articles that have been manufactured or refine for sale, like devices, medicine, food, well-defined services.
    \item \texttt{title} for title of movies, books, TV shows, songs, etc. Titles can be sentences and they usually refer to media, which can be considered a fine-grained version of product. 
    \item \texttt{event} for situations or scenarios that gather people for a specific purpose such as concerts, competitions, conferences, award events, etc. Events do not consider holidays. 
    \item \texttt{time} for months, days of the week, seasons, holidays and dates that happen periodically, which are not events (e.g., Christmas). It excludes hours, minutes, and seconds.
    \item \texttt{other} for any other named entity that does not fit in the previous categories.
\end{itemize}

% \section{Data Distribution}
% \label{app:data-distribution}

\end{document}